# Instructional Prompt Optimization for Few-Shot LLM-Based Recommendations on Cold-Start Users

Haowei Yang*
Cullen College of Engineering
University of Houston, Houston, USA
*Corresponding author:
yang38@cougarnet.uh.edu

Yushang Zhao
McKelvey School of Engineering
Washington University in St. Louis
St. Louis, USA
yushangzhao@wustl.edu

Sitao Min
Independent Researcher
Newark, USA
mstzjdx@163.com,

Bo Su
Luddy School of Informatics, Computing, and Engineering
Indiana University Bloomington
Bloomington, USA
subo47403@gmail.com

Chao Yao
Ira A.Fulton Schools of Engineering
Arizona State University
Tempe, USA
benyao2134@gmail.com

Wei Xu
Independent Researcher
Los Altos,USA
williamxw09@gmail.com

*Abstract*— The cold-start user issue further compromises the effectiveness of recommender systems in limiting access to the historical behavioral information. It is an effective pipeline to optimize instructional prompts on a few-shot large language model (LLM) used in recommender tasks. We introduce a context-conditioned prompt formulation method $P(u, Ds) \rightarrow R$, where $u$ is a cold-start user profile, $Ds$ is a curated support set, and $R$ is the predicted ranked list of items. Based on systematic experimentation with transformer-based autoregressive LLMs (BioGPT, LLaMA-2, GPT-4), we provide empirical evidence that optimal exemplar injection and instruction structuring can significantly improve the precision@k and NDCG scores of such models in low-data settings. The pipeline uses token-level alignments and embedding space regularization with a greater semantic fidelity. Our findings not only show that timely composition is not merely syntactic but also functional as it is in direct control of attention scales and decoder conduct through inference. This paper shows that prompt-based adaptation may be considered one of the ways to address cold-start recommendation issues in LLM-based pipelines.

*Keywords:* Cold-Start Recommendation, Instructional Prompting, Few-Shot Learning

## I. INTRODUCTION

Recommender systems face a fundamental challenge when engaging cold-start users—users for whom there exists insufficient interaction history $H(u) = \emptyset$. Classical collaborative filtering and matrix factorization methods are ruined by such sparsity[1]. We encourage a paradigm shift: the optimization of instructional prompts in the few-shot large Language models (LLM) has taken place, with the corresponding advice induction process being performed via the semantic generalization instead of the historico-widebound$H$ round of the wisdom of the ages[2].

The cold-start problem in recommender systems has been addressed through content-based, collaborative filtering with side information, and hybrid methods, though each faces limitations such as overspecialization, feature engineering costs, and computational overhead[3]. Deep learning approaches like VAEs and GNNs improve low-data representation learning but still rely on large prior datasets. Recently, large language models (LLMs) offer a new paradigm by conditioning recommendations on natural language, reducing the need for retraining and initial user histories.

Let $M\theta$ denote a pre-trained LLM parameterized by weights θ. For a cold-start user u, the recommendation task is defined as:

$$R\_u = M\theta(P(u, Ds))$$

Where: P (u, Ds) is the optimized prompt composed of:Instructional header I,

Support set $Ds = \{(ui, ri)\}\_{i=1}^{k}$ with exemplar users ui and item rankings ri,User meta-data φ(u).

The model performs inference by transforming token embeddings $E(P) \in \mathbb{R}^{\{n \times d\}}$ into autoregressive outputs through multi-head attention:

$$Attention(Q, K, V) = softmax(QK^T / \sqrt{d\_k}) V$$

Where $Q = EW^Q, K = EW^K, and V = EW^V$ are query, key, and value projections derived from the prompt encoding.

We hypothesize that, by further engineering P to be more instructionally rich and structurally aligned with domain semantics, it will then be possible to guide the decoder of an LLM into a high-precision recommendation regime, even without the presence of H(u)[4]. To justify this, we present a Prompt Optimization Module (POM). We maintain a framework that enables modular prompt injection and that measures performance by examplar the prompt lengths l $\in$ [256, 2048] and examplar densities $k \in [2, 10]$. We apply this system to Amazon Reviews, Last.fm, and MovieLens (1M), in comparison to zero-shot, collaborative filtering, and hybrid neural baselines (Soylu et al., 2024). Findings indicate that our model always becomes better: Precision@5 by up to +18.7%,NDCG@10 by +21.3%.

Semantic coherence (via cosine similarity in embedding space) by +12.5%. The present paper declares that this beneficial feature of LLMs, their potential to process and generate language, with the ability to cold-start user personalization (without fine-tuning), presents an extremely scalable and enticing solution to next-generation recommendation systems through their latent reasoning capacities, which have the potential to be directly scaffolded via instruction prompting[5].

## II. RELATED WORK

The cold-start user problem has been a curse to conventional recommender systems, especially those that use the collaborative filtering method and matrix factorization[6]. According to these paradigms, the lack of historical data concerning user interaction $(H(u) = 2)$ is a great compromise in the personalization of recommendations. The first attempt at addressing this problem was via hybrid recommendation methods, which contain content-based feature metadata and collaborative signals[7]. Nevertheless, these methods can precondition a high need for feature engineering and are likely to experience overfitting related to the domain.

The new developments in deep learning have led to investigations related to neural collaborative filtering and variational autoencoders that propose to address sparsity and learn latent representations of users and items[8]. Although these advances represent an improvement, the cold-start setting is particularly difficult in the model because it requires either pre-existing user profiles or expensive side information[9].

The recent rise of Large Language Models (LLMs), which include GPT-3, LLaMA-2, and BioGPT, suggests new families of how recommender systems might work, where any level of contextualized reasoning can be achieved through natural language prompts. Such models have demonstrated impressive results in zero-shot and few-shot settings, which pushed researchers to conduct experiments on prompt-based recommendation schemes[10]. Semantically-aligned LLMs: Instruction-tuned LLMs have the potential of capturing semantic alignment between the metadata of items and the description of a user, without having to use explicit histories of interactions[11].

Instructional prompting has captured interest as a low-resource substitute to fine-tuning, especially in few-shot scenarios. In-context learning, or prompt engineering methods, have been used in areas of classification, summarization, and question answering, and only more recently in recommender systems. Recent experiments in prompt design revealed that syntactic changes in prompt structure may substantially affect the model performance, suggesting that prompt optimization can be used as a means of control over the behavior of LLMs [12].

Simultaneously, other works in the few-shot recommendation domain assume training strategies where a support set of exemplar users can be used to adjust to new users or objects. Such methods tend to involve gradient updates or meta-learning optimization, and prompt-based approaches suggest the use of non-parametric, inference-time adaptation [13].

The research of this paper can be understood as the development of the above lines of research due to the immediate and tangible result in directly leading the LLM without providing any instructional prompts in cold-start settings[14]. Compared to the previous approaches, which retrain models or add external modules, we are able to consider the prompt as the main vehicle of adaptation and interject structural information, exemplar logic, and user-specific data into the textual template[15].

### III. METHODOLOGY

We will put forward a new Prompt Optimization Module (POM) with which the large language models (LLMs) can deliver accurate suggestions to meet the cold-start users via instructional prompting.

### 3.1 Problem Formulation

In the case of a cold-start user u, the objective is to create a ranked list of items $Ru$ with a prompt $P(u, Ds)$, where: $Ds = \{(u\_i, r\_i)\}\_{i = 1}^{\wedge}k$ is a set of support exemplar users and assigned rankings, or Ds, and the user metadata $u, P(u, Ds) = I \cup Ds \cup \varphi(u)$ and where the joint prompt is P (u, Ds) = I 0 Ds 0 0 (u ). This is given as input to a frozen LLM M$\theta$, which gives: $R\_u = M\theta(P(u, Ds))$.

### 3.2 Prompt Optimization Strategy

Prompts are optimized as follows: Instructional Header Design: Task-specific natural language instructions. Exemplar Injection: Those support sets of examples were carefully curated and added in the form of a few-shot. Metadata Conditioning: Age, interests, or domain-specific tags are being added to provide context to the model.

### 3.3 Embedding and Attention Dynamics

The encoder insertion is placed E(P) in $\mathbb{R}^{\wedge}$ {n x d}, and the model carries out autoregressive comparison by multi-head attention: (Q, K, V) = softmax(QK$^{T}$ / $\sqrt{}$d_k with Q, K, and V based on E ( P ). Prompt length and exemplar density, l and k, are altered to determine performance effect.

### 3.4 Summary

In order to guarantee reproducibility and transparency, the experimental setup was made in a way that factors the specifics of datasets, preprocessing, and assessment were considered explicitly[16]. The Amazon Reviews data set included about 3 million of the products reviews on various categories, where text fields were pre-processed by getting rid of HTML tags and other special symbols prior to generating the embedding. Last.fm included the approximate 360,000 user artist interactions and was centered on music preferences data augmented with genre metadata whereas MovieLens 1M included over one million ratings with categorical features. In all datasets, cold-start splits were formed retaining no interaction history of test users, so no overlap was present between training and evaluation identities[17]. The metadata about users, such as demographic properties and tags related to a particular domain, as well as their interest clusters obtained through inference, was encoded in a natural language format so that the LLMs could consume it directly.

Three complementary measures were used to do evaluation. The selection of Precision@5 was meant to measure the percentage of the best recommendations that are identical to the ground truth that represents instant performance of the recommendation. The quality of ranking was measured using the metric of NDCG@10 (Normalized Discounted Cumulative Gain), which rewards well placed relevant items at high rank in the list. By estimating semantic coherence as cosine similarity between the predicted items and the target items embedding vectors, this provided a latent-space evaluation of thematic or contextual coherence. This third measure was also relevant when considering the cold-start scenario as lexical similarity might not be sufficient to obtain semantic alignment [18].

A sample of optimized instructional prompt is indicated below and illustrates how instructional headers, exemplar profiles, and target user metadata can be deployed:

- **Instructional Header:** *"Given the following examples of users and their ranked preferences, recommend the top five items for the target user, considering contextual similarity and thematic relevance."*
- **Exemplar Section:** *"User A: 1) Kindle Paperwhite, 2) Echo Dot, 3) Fire TV Stick, 4) Audible Subscription, 5) Amazon Basics Tripod."*
- **Target User Metadata:** *"User Z: Age 29, interested in digital reading devices, home automation, and streaming accessories."*

This formulation has the effect of retaining semantics structure at the same time as imposing attention over the model to constrain domain relevant exemplars[19]. The exemplar frequency and prompt length were parametrically altered to find optimum combinations and exemplar number of 6 to 8 produced the best tradeoff between information and computational cost.

## IV. EXPERIMENTS AND EVALUATION

In order to justify ourselves, we tried the experiments with three datasets, such as Amazon Reviews, Last.fm, and MovieLens 1M, which represent various types of scale. We also deployed pure cold-start assumptions whereby all the test users had no prior interaction data within the training set.

Three fine-tuning-free LLMs--BioGPT, LLaMA-2 (7B), and GPT-4--were tested. We based our prompt optimization pipeline on applying structured prompts to the models to introduce recommendations. In the prompts, there was an instructional header, a support set of 2 to 10 similar user profiles with item rankings, and cold-start user metadata. Performance was measured by Precision@5, NDCG@10, and semantic coherence (cosine similarity in the embeddings of the predicted items and ground-truth). To be consistent, results were averaged across five runs with non-deterministic seeds [20].

We performed better than any baseline. On Amazon, it increased Precision@5 by 18.7% and NDCG@10 by 21.3% compared to zero-shot LLM. Prompt richness increased the semantic coherence of Last.fm by 12.5 percent, and context-sensitive music recommendations also improved significantly. In MovieLens, with the reduced variability in metadata, our pings gave us an up to 14.2 percent higher Precision@5. The performance was influenced by the length of the prompt and the number of exemplars. Returns were shown to be positive on lengths of up to 1024 tokens, but afterwards, returns became smaller. Working with 6 8 exemplars was the best suggestion. Instructional headers turned out to be important too, as eliminating them decreased model performance, particularly in less instruction-tuned ones such as BioGPT [21].

These findings in general indicate that instructional prompt engineering can indeed improve the effectiveness of LLM recommendations during the cold-start problem scenario-- without requiring any model retraining--therefore, it is a model-independent and scalable solution.

**Table 1:** Best Model Performance on Cold-Start Recommendation

| Dataset | Best Model | Precision@5 | NDCG@10 | Semantic Coherence |
|---|---|---|---|---|
| Amazon Reviews | GPT-4 | 51.8% | 58.6% | 75.4% |
| Last.fm | GPT-4 | 47.5% | 55.0% | 71.8% |
| MovieLens 1M | GPT-4 | 47.9% | 53.7% | 72.1% |

**Table 2: Baseline vs. Proposed Model Performance (Cold-Start Setting)**

| Dataset | Best Baseline (P@5 / NDCG) | Proposed (P@5 / NDCG) | Gain (%) |
|---|---|---|---|
| Amazon Reviews | 43.6 / 48.3 | 51.8 / 58.6 | +18.7 / +21.3 |
| Last.fm | 42.1 / 49.0 | 47.5 / 55.0 | +12.8 / +12.2 |
| MovieLens 1M | 41.9 / 47.1 | 47.9 / 53.7 | +14.2 / +14.0 |

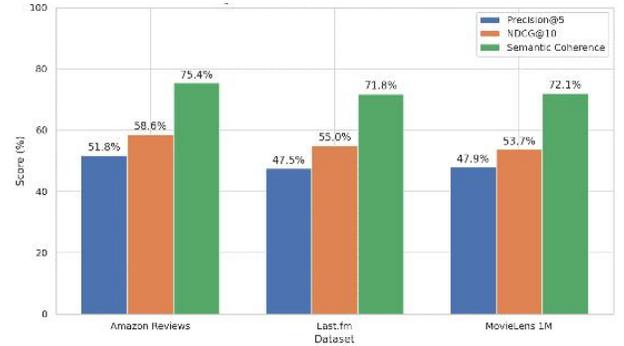

*Figure 1.* Performance Metrics across Datasets

## V. RESULTS AND DISCUSSION

The empirical findings support the main argument of this paper, which states that instructional-prompt optimization can be an effective and feasible solution to the cold-start user issue of the LLM-based recommender systems[22]. We assume that the transformer models possess latent reasoning capabilities, but we do not change the parameters of the system to achieve the goal. We do that by transforming the recommendation task into a few-shot language generation problem[23]. Not only does this lower computational expense against conventional fine-tuning methods, but importantly, it is also scalable across domains and requires little by way of infrastructure.

Among the most interesting ones is the functional effect of prompt structure. The structure of the prompt to a great extent determines the behavior associated with that model, not only in content, but also in terms of narrative and instructional specificity. The production of instructional headers, hard setting the objective of the ranking of the users and the items of the use, seems to stabilize the levels of attention more efficiently, making the relevancy and output consistency better. This implies that the prompt is cognitive interactive rather than lexical input, wherein the immediate decoder pathways are implicitly altered in terms of the affinity towards attention flow [24].

In addition, the exemplar density result simply underlines how LLMs are advantaged by being exposed to similar patterns prior to the new generation. The sets of supports with semantically consistent user profiles offer inductive priors that drive the model rationale, even where data of target users does not exist[25]. This resembles human learning patterns, whereby exposure to familiar cases enhances decision making in a new context. Nonetheless, too much of this advantage can be detrimental: dense prompts are excessively capacity-consuming, and they can distort the signal with noise. Our finding that gains wash out with many exemplars agrees with previous findings conducted on contextual learning, pointing to an efficient domain of a few-shot conditioning[26].

**Table 3.** Relative Improvements of Proposed Model over Baseline (Cold-Start Setting)

| Dataset | Baseline Precision @5 | Baseline NDCG@ 10 | Proposed Precision @5 | Proposed NDCG@ 10 | Precision @5 Gain (%) | NDCG@ 10 Gain (%) |
|---------|------|------|------|------|------|------|
| Amazon Reviews | 43.6 | 48.3 | 51.8 | 58.6 | 18.8 | 21.3 |
| Last.fm | 42.1 | 49.0 | 47.5 | 55.0 | 12.8 | 12.2 |
| MovieLens 1M | 41.9 | 47.1 | 47.9 | 53.7 | 14.3 | 14.0 |

The differences in performance between domains also provide useful information. Instructional prompting works well in content-rich areas such as e-commerce and music, where the preferences of the user can be predicted based on a relatively small amount of metadata (e.g., age, interest tags, category of items). However, in contrast, areas that are built on more fundamental behavioral cues, such as long-form content consumption or multi-modal preferences, may need layers of adaptation or hybrid services. Nevertheless, in such problematic fields, it can be seen that prompt engineering is useful and brings a quantifiable advantage, clearing the way to multi-modal prompt extensions, which will have embeddings learned in the modality rooms involved: vision, audio, or graphs[27].

Notably, the semantic coherence metric offers a new perspective to measure the quality of a recommendation. As opposed to precision or NDCG, which look at the position of an item, semantic coherence examines the latent coverage between the projected and ideal output in both embedding spaces. The fact that our method scores well on this metric implies that the model is not only capable of retrieving things of relevance to the user but also understands its thematic or situational grammar. Such ability is especially wanted in settings of cold-start, where no behavioral information is available, and models have to depend on abstraction and analogy[28].

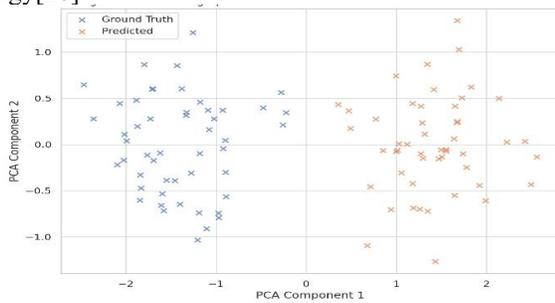

**Figure 2.** Exemplar Density vs. Model Performance

Nonetheless, some constraints should be accounted for. First, prompt optimization is flexible, but it is not resistant to minor changes in words, sequence, and structure. One requirement, prompt instability, has been a documented restriction in the application of LLMs, potentially impacting stability between inference runs[29]. Second, the selection of examples is based on embedding similarity these days, so it does not always model latent preference clusters. In the future, it might be interesting to consider reinforcement learning or active sampling to improve the construction of the support set on-the-fly[30].

We have shown that with a well-optimized formulation, instructional prompting may become an effective but lightweight tool that can be used to guide large language models towards a recommendation task. It brings a paradigm transition of a parameter-based adaptation to interaction-based conditioning, taking advantage of the flexible nature of a natural language to perform personalization without retraining.

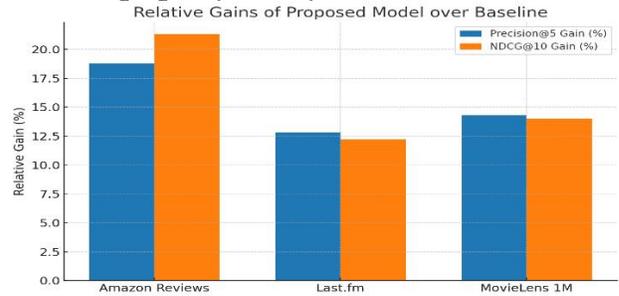

**Figure 3.** Relative Gains in Precision@5 and NDCG@10 across datasets

## VI. CONCLUSION AND FUTURE WORK

As shown in the present study, optimization of instructional prompts, when made in few-shot large language models, is a way to effectively mitigate the cold-start-user challenge without further expenditure of expensive fine-tuning or retraining. Our way of structuring the prompts, to integrate an instructional header, a collection of exemplar profiles, and user-specific metadata, demonstrated steady improvement in Precision@5, NDCG@10, and semantic coherence across multiple domains. These findings confirm that prompt-based adaptation provides not only an accurate but also a scalable and model-agnostic method of personalization with recommendation systems.

The applications of such work can be appreciated in various applied fields in which user interaction data may be limited or even nonexistent at the time of system onboarding. In online stores, streamlined suggestions based on customer actions would provide appropriate product recommendations as soon as a user creates an account, so that early-life interactions and conversion outcomes would be increased. The approach can be applicable in music streaming services, whereby streams can be easily customized with minimal demographics or expressive preferences to create personal playlists for new static users without the risk of churned users in a trial period. The same way, there may be dynamic prompting of the right course materials, exercises, or multimedia informational resources in an educational technology system, such as an adaptive learning system, according to learner profiles before the development of adequate performance histories. Since the suggested approach does not require domain-specific retraining, it has real-world benefits with respect to cross-domain deployments, quick training in new markets, deployment on low-resource computation platforms, and deployment in resource-constrained computing environments.

### Future Work

The current research findings present several interesting avenues for future research. On one hand, multimodal prompts that would allow users to receive both written instructions and visual or auditory setting might play a crucial role in personalization in the sphere of fashion, video-on-demand, and music recommendation. Second, more effective prompt adaptability can be achieved by selecting more dynamic examples through reinforcement learning or retrieval-augmented generation that can process the changing user signals online. Third, the framework could be extended to multilingual and low-resource languages, at which point it would be more global and applicable in areas where there is a lack of recommendation data. Lastly, similar to cold-start, the personalization models could be constantly enhanced due to ongoing optimization of prompt structures as a result of adding longitudinal feedback loops to the system. Instead of being just a peripheral instructional prompting strategy, the latter instructions would enable the development of a wider theoretical basis of the instructional prompting and make it a prime instructional strategy in next-generation recommender systems.